\documentclass{bmvc2k}

\pdfoutput=1

\usepackage{times}
\usepackage{epsfig}
\usepackage{graphicx}
\usepackage{amsmath}
\usepackage{amssymb}
\usepackage{adjustbox}
\usepackage{mathtools}
\newcommand{\norm}[1]{\left\lVert#1\right\rVert}

\title{Shape Generation using Spatially Partitioned Point Clouds}

\addauthor{Matheus Gadelha}{mgadelha@cs.umass.edu}{1}
\addauthor{Subhransu Maji}{smaji@cs.umass.edu}{1}
\addauthor{Rui Wang}{rui.wang@cs.umass.edu}{1}

\addinstitution{
 University of Massachusetts, Amherst\\
 Amherst, MA, USA
}

\runninghead{Gadelha et al.}{Shape Generation using Spatially Partitioned Point Clouds}

\def\eg{\emph{e.g}\bmvaOneDot}

\def\etal{\emph{et al}\bmvaOneDot}

\begin{document}

\maketitle

\begin{abstract}

We propose a method to generate 3D shapes using point clouds.
Given a point-cloud representation of a 3D shape, our method builds a \emph{kd-tree} to spatially partition the points. This orders them consistently across all shapes, resulting in reasonably good correspondences across all shapes. 
We then use PCA analysis to derive a linear shape basis across the spatially partitioned points, and optimize the point ordering by iteratively minimizing the PCA reconstruction error. Even with the spatial sorting, the point clouds are inherently noisy and the resulting distribution over the shape coefficients can be highly multi-modal. We propose to use the expressive power of neural networks to learn a distribution over the shape coefficients in a generative-adversarial framework. 
Compared to 3D shape generative models trained on voxel-representations, our point-based method is considerably more light-weight and scalable, with little loss of quality. It also outperforms simpler linear factor models such as Probabilistic PCA, both qualitatively and quantitatively, on a number of categories from the ShapeNet dataset. Furthermore, our method can easily incorporate other point attributes such as normal and color information, an additional advantage over voxel-based representations.
\end{abstract}

\def\point{P}
\def\matrix{P}
\def\npoints{N}
\def\nshapes{S}
\def\nbasis{B}

\vspace{-12pt}
\section{Introduction}\label{sec:intro}
The choice of representation is a critical for learning a good generative model of 3D shapes. Voxel-based representations that discretize the geometric occupancy into a fixed resolution 3D grid offers compelling advantages since convolutional operations can be applied. 
However, they scale poorly with the resolution of the grid and are also wasteful since the geometry of most 3D shapes lies on their surfaces, resulting in a voxel grid that's mostly empty, especially at high resolutions. 
Surface-based representations such as triangle meshes and point clouds are more efficient for capturing surface geometry, but these representations are inherently unstructured -- because there is no natural ordering of the points, they are better expressed as an unordered \emph{set}. Consequently, unlike \emph{ordered} representations, they are cannot be easily generated using existing deep convolutional architectures. 
The exception is when the points are in perfect correspondence across shapes, in which case a linear shape basis can be effective (\eg, for faces or human bodies). However estimating accurate global correspondences is difficult and even poorly defined for categories such as chairs that have complex and varying geometry. Thus generating 3D shapes as point clouds remains a challenge.

\begin{figure}
\centering
\includegraphics[width=0.8\linewidth]{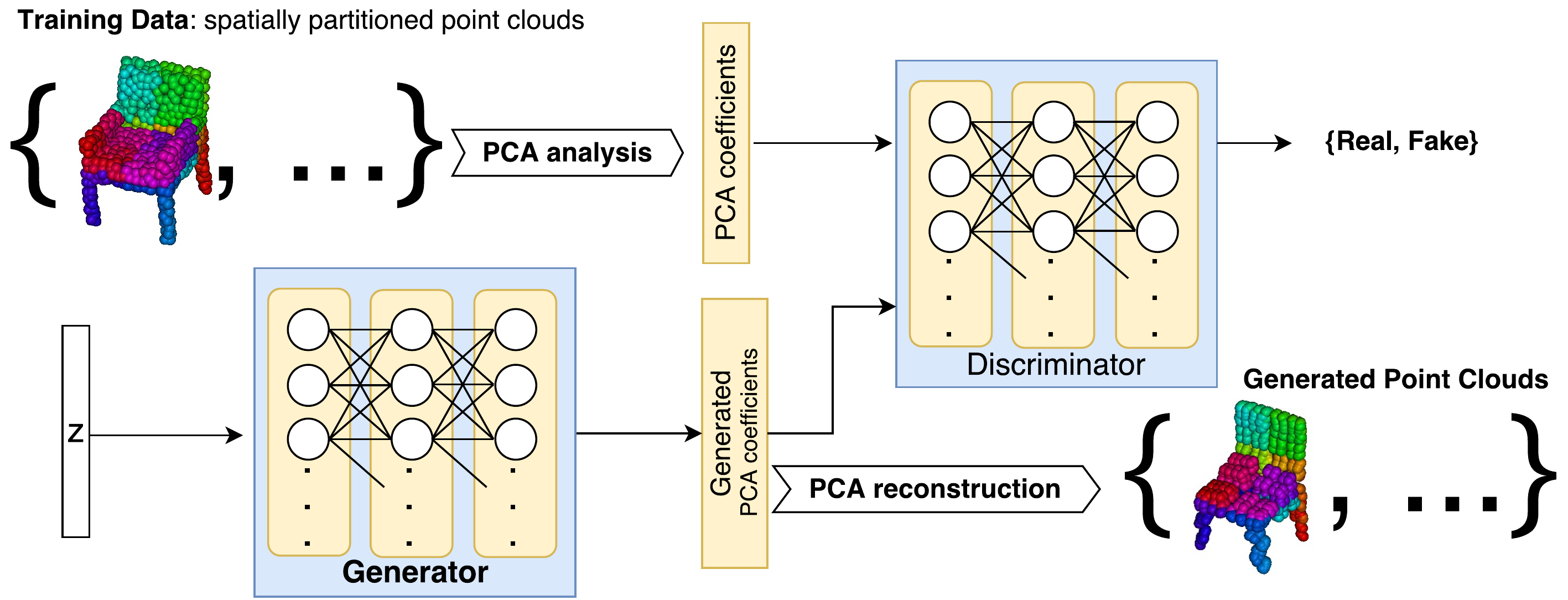}
\caption{\small \label{fig:arch} Our network architecture for generating 3D shapes using spatially partitioned point clouds. We perform PCA analysis on the training data to drive a shape basis and associated shape coefficients. We then train a GAN to learn the multi-modal distribution over the coefficients. The generated coefficients are combined with the shape basis to produce the output point clouds.}
\vspace{-12pt}
\end{figure}

We propose a new method for learning a generative model for 3D shapes represented as point clouds. Figure~\ref{fig:arch} illustrates our network architecture. 
The key idea is to use a space-partitioning data structure, such as a \emph{kd-tree}, to approximately order the points. 
Unlike a voxel-grid occupancy representation, the kd-tree representation scales linearly with the number of points on the surface and can adapt to the geometry of the model.
Moreover one can easily incorporate other point attributes such as surface normal, color, and texture coordinates into this representation, making it possible to generate new shapes that automatically include these information.
We learn a shape basis over the ordered point clouds using PCA analysis of the shape coordinates. The point ordering of each shape is then optimized by iteratively minimizing the PCA reconstruction error. Subsequently a new shape basis can be learned on the reordered points.
If the alignments induced by the kd-tree sorting was perfect, the distribution of the coefficients would be simple.
Indeed this is the assumption behind generative models such as Probabilistic PCA~\cite{tipping1999probabilistic} that models the distributions of coefficients using independent Gaussians. 
However, imperfect alignment can lead to a multi-modal and heavy-tailed distribution over the coefficients.
To address this issue, we propose to leverage the expressive power of neural networks and employ a Generative Adversarial Network (GAN)~\cite{goodfellow2014generative} to learn the distribution over the shape coefficients.
Unlike other non-parametric distributions such as a mixture of Gaussians, the GAN linearizes the distribution of shapes and allows interpolation between them using arithmetic operations.
At the same time our method remains light-weight and scalable, since most shape categories can be well represented with a hundred basis coefficients.

We compare the proposed generative model to a 3D-GAN approach of Wu~\etal~\cite{wu2016learning} that learns a convolutional architecture over a voxel-representation of 3D shapes. In addition we compare to a Probabilistic PCA (PPCA) baseline using the same point-cloud representation. Experiments on several categories in the ShapeNet dataset show that the proposed approach outperforms PPCA and 3D-GAN, quantitatively and qualitatively. Compared to the 3D-GANs our models are an order-of-magnitude faster and smaller. We then present several experiments evaluating the role of the kd-tree on the quality of the generated shapes. We also show that a 1D-convolutional GAN trained on the ordered list of point coordinates produces samples of reasonable quality, suggesting that the kd-tree ordering plays a key role.

\vspace{-12pt}
\section{Related Work}\label{sec:related}

\noindent \textbf{Generative models for 3D shapes.} Recently, Wu~\etal in~\cite{wu2016learning} proposed a generative model of 3D shapes represented by voxels, using a variant of GAN adapted to 3D convolutions. Two other works are also related. Rezende~\etal~\cite{rezende2016unsupervised} show results for 3D shape completion for simple shapes when views are provided, but require the viewpoints to be known and the generative models are trained on 3D data. Yan~\etal in~\cite{yan2016perspective} learn a mapping from an image to 3D using multiple projections of the 3D shape from known viewpoints (similar to a visual-hull
technique). However, these models operate on a voxel representation of 3D shape, which is difficult to scale up to higher resolution. The network also contains a large number of parameters, which are difficult and take a long time to train. Our method uses spatially partitioned point cloud to represent each shape. It is considerably more lightweight and easy to scale up. In addition, by using a linear shape basis, our network is small hence much easier and faster to train. Through experiments we show that the benefits of this lightweight approach come with no loss of quality compared to previous work. Several recent techniques~\cite{tatarchenko2017octree,Riegler2017CVPR} have explored multi-resolution voxel representations such as \emph{octrees}~\cite{meagher1982geometric} to improve their memory footprint at the expense of additional book keeping. But it remains unclear if 3D-GANs can generate high-resolution sparse outputs.

\vspace{8pt}
\noindent \textbf{Learning a 3D shape bases using point-to-point correspondence.} Another line of work aims to learn a shape basis from data assuming a global alignment of point clouds across models. 
Blanz and Vetter in~\cite{blanz1999morphable} popularized the 3D morphable models for faces which are learned by a PCA analysis of the point clouds across a set of faces with known correspondences. 
The same idea has also been applied to human bodies~\cite{allen2003space}, and other deformable categories~\cite{kar2015category}. 
However, establishing the point-to-point correspondence between 3D shapes is a challenging problem. 
Techniques are based on global rigid or non-rigid pairwise alignment (\eg, \cite{besl1992method,chen1992object,bronstein2010gromov}), learning feature descriptors for matching (\eg, techniques in this survey~\cite{van2011survey}), or fitting a parametric model to each instance (\eg,~\cite{cashman2013shape,prasad2010finding}).
Some techniques improve pairwise correspondence by enforcing cycle-consistency across instances~\cite{huang2013consistent}. 
However, none of these techniques provide consistent global correspondences for shapes with varying and complex structures (\eg, chairs and airplanes). 
Our method uses spatial sorting based on a kd-tree structure. It is a fast and lightweight approximation to the correspondence problem. However, unlike alignment-based approaches, one drawback of the kd-tree sorting is that it's not robust to rotations of the model instances. This is also a drawback of the voxel-based representations.
The ShapeNet dataset~\cite{chang2015shapenet} used in our experiments already contains objects that are consistently oriented, but otherwise one can apply automatic techniques (\eg,~\cite{Su_2015_ICCV}) for viewpoint estimation to achieve this.

\vspace{8pt}
\noindent \textbf{Representing shapes as sets.} Another direction is to use loss functions over \emph{sets} such as Chamfer, Hausdroff, or Earth Mover's Distance (EMD) to estimate similarity between models. The recent work of Fan~\etal~\cite{fan2016point} explores this direction and trains a neural network to generate points from a single image by minimizing the EMD between the generated points and the model points. To apply this approach for shape generation one requires the evaluation of the loss of a generated shape with respect to a distribution of target shapes. While this can be approximated by computing statistics of EMD distance of the generated shape to all shapes in the training data, this is highly inefficient since EMD computation alone scales cubically with the number of points. 
Thus training neural architectures to generate and evaluate loss functions over sets efficiently remains an open problem. 
The approximate ordering induced by the kd-tree allows efficient matrix operations on the \emph{ordered} vector of point coordinates for training shape generators and discriminators.

\section{Method}\label{sec:method}

\begin{figure}
\centering
	\includegraphics[width=0.49\linewidth]{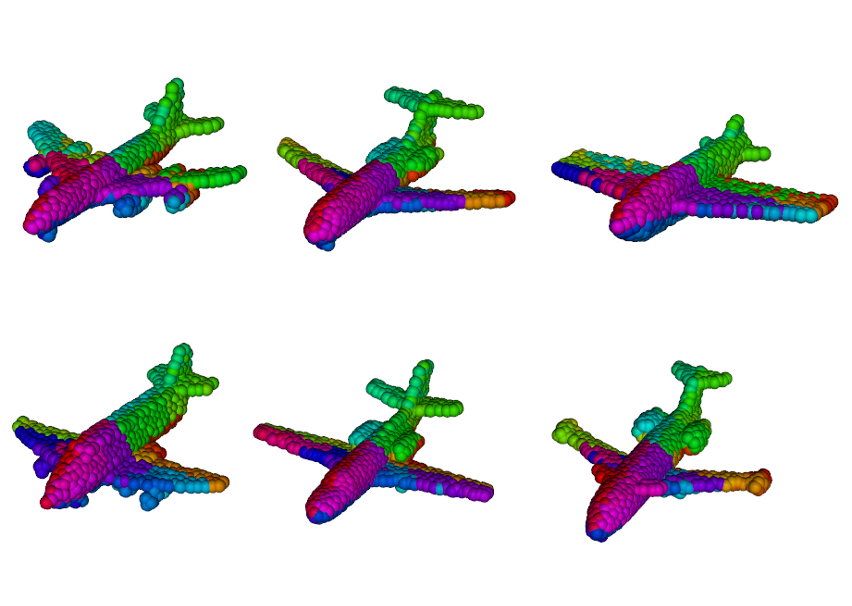}
	\includegraphics[height=1.4in]{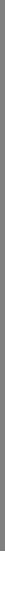}
	\includegraphics[width=0.49\linewidth]{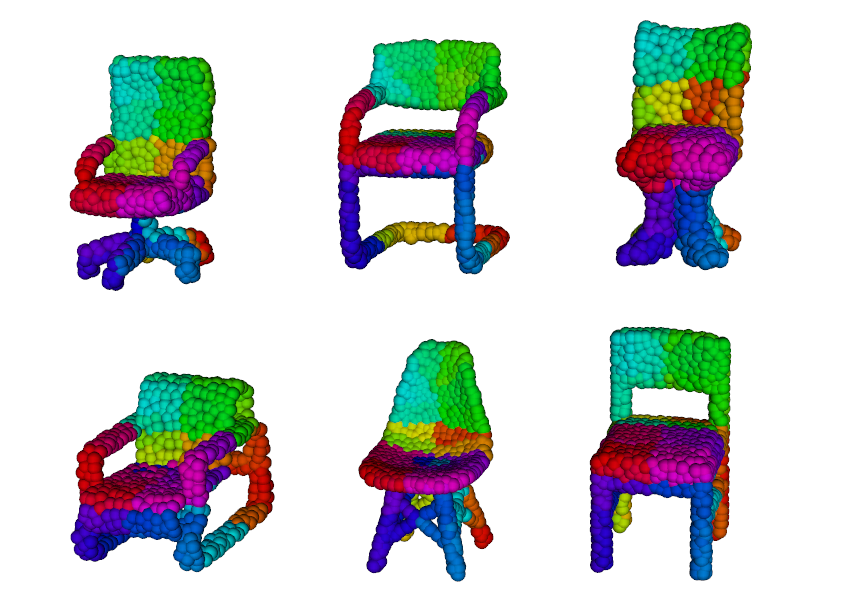}
	\vspace{-12pt}
	\caption{\small \label{fig:point_sorting} Visualization of spatially partitioned points for six training shapes from each category. Every point is colored by its index in the sorted order. This shows that the kd-tree sorting leads to reasonably good correspondences between points across all shapes.}
	\vspace{-12pt} 
%
\end{figure}

This section explains our method. To begin, we sample each training 3D shape using Poisson Disk sampling~\cite{Bowers:2010:PPD} to generate a consistent number of evenly distributed points per shape. We typically sample each shape with 1K points, and this can be easily increased or decreased based on actual need. We then build a kd-tree data structure for each point cloud to spatially partition the points and order them consistently across all shapes. Next, we compute the PCA bases using all the point data. 
Finally, we train a GAN on the shape coefficients to learn the multi-modal distribution of these coefficients and use it to generate new shapes.

\vspace{12pt}
\noindent \textbf{Spatially partitioned point cloud.} We use $\{\point_i^s\}$ to represent a point cloud where $i$ is the point index and $s$ is the shape index. By default the point data $\point$ includes the $x,y,z$ coordinates of a point, but can include additional attributes such as normal and color etc. We assume each point cloud is centered at the origin and the bounding box is normalized so that the longest dimension spans [-0.5, 0.5]. For each point cloud we build a kd-tree by the following procedure: we start by sorting the entire point cloud along the $x$-axis, and split it in half, resulting in a left subset and a right subset; we then recursively split each of the two subsets, but this time along the $y$-axis; then along $z$-axis, and so on. Basically it's a recursive splitting process where the splitting axis alternates between $x$, $y$, and $z$. The splitting axes can also be chosen in other ways (such as using the longest dimension at each split) to optimize the kd-tree building, but it needs to be consistent across all point clouds.

The kd-tree building naturally sorts the point cloud spatially, and is consistent across all shapes. For example, if we pick the first point from each sorted point cloud, they all have the same spatial relationship to the rest of the points. As a result, this establishes reasonably good correspondences among the point clouds. Figure~\ref{fig:point_sorting} shows an illustration.

\vspace{12pt}
\noindent \textbf{Computing PCA bases.} We use PCA analysis to derive a linear shape basis on the spatially partitioned point clouds. To begin, we construct a matrix $\matrix$ that consists of the concatenated $x,y,z$ coordinates of each point cloud and all shapes in a given category. The dimensionality of the matrix is $3\,\npoints\times\nshapes$ where $\npoints$ is the number of points in each shape, and $\nshapes$ is the number of shapes. We then perform a PCA on the matrix: $\matrix=U \Sigma V$, resulting in a linear shape basis $U$. Thanks to point sorting using kd-tree, a small basis set is sufficient to well represent all shapes in a category. We use $\nbasis$ to represent the size of the shape basis, and by default choose $\nbasis=100$, which has worked well for all ShapeNet categories we experimented with. The choice of $\nbasis$ can be observed from the rapid dropping of singular values $\Sigma$ following the PCA analysis. Without a good spatial sorting method, it would require a significantly larger basis set to accurately represent all shapes.

To include other point attributes, such as normal, we can concatenate these attributes with the $x,y,z$ coordinates. For example, a matrix that consists of both point and normal data would be $6\,\npoints\times\nshapes$ in size. We suitable increase the basis size (e.g. by choosing $\nbasis=200$) to accommodate the additional data. The rest of the PCA analysis is performed the same way.

\begin{figure}
\centering
	\includegraphics[width=0.5\linewidth]{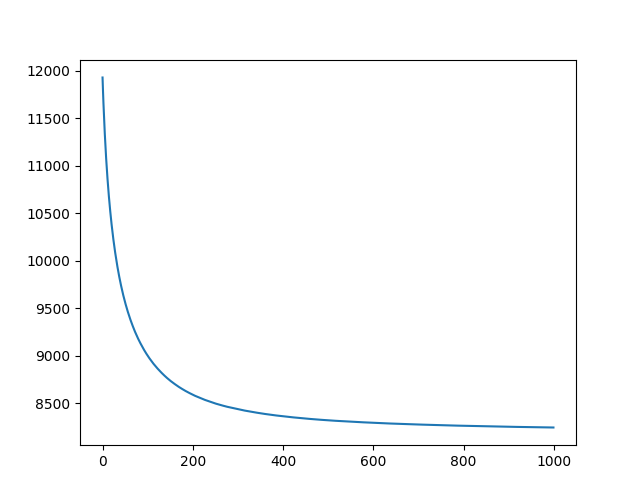}
	\vspace{-12pt}
	\caption{\small \label{fig:sort_error} Decay of PCA reconstruction error following $I=1000$ iterations of the point optimiation procedure. The vertical axis represents the PCA reconstruction error and the horizontal axis represents the number of iterations.}
	\vspace{-12pt} 
\end{figure}

\vspace{12pt}
\noindent \textbf{Optimizing point ordering.} While sorting using the kd-tree creates good initial correspondences between points, 
the point ordering can be further optimized by iteratively reducing the PCA reconstruction error through the following procedure. 
For shape's point cloud $\{\point_i^s\}$ (where $s$ is the shape index and $i$ is the point index), we perform random swapping $K$ times. 
Specifically, we first randomly select a pair of points $\langle \point_i^s, \point_j^s \rangle$ and make them candidates for swapping. 
If the resutling PCA reconstruction error is reduced, we swap the two points. 
This is repeated $K$ times. 
The reconstruction error of a vectorized point cloud $P^s$ using a basis $U$ is computed as follows:
\begin{equation}\label{eqn:recerror}
	\mathcal{L}_{rec}(P^s, U)  = \norm{(P^s - \mu)^T U^T U + \mu - P^s}_2^2,
\end{equation}
where $\mu = \frac{1}{|\mathcal{D}|}\sum_{s \in \mathcal{D}} P^s$.
After every shape is processed, we then re-compute a new PCA basis using the optimized point ordering. 
Finally, the whole procedure is repeated $I$ iterations. 
In our experiments, we have chosen to use $K=10^4, I=10^3$. 
Figure \ref{fig:sort_error} shows the decay of reconstruction error during the optimization procedure. 
The shapes used in this figure are chair models from the ShapeNet dataset. 
Experiments show that the point optimization improves the results both qualitatively and quantitatively.

\vspace{12pt}
\noindent \textbf{Learning shape coefficients using GAN.} Our method employs a GAN to learn the distribution over the shape coefficients. 
Following the PCA analysis step, the matrix $V$ captures the coefficients for all training shapes, i.e. the projections of each point cloud onto the PCA basis. It provides a compact and yet accurate approximation of the 3D shapes. Therefore our generative model only needs to learn to generate the shape coefficients. 
Since the dimensionality of the shape basis ($\nbasis=100$) is much smaller (than the number of points on each shape), we can train a GAN to learn the distribution of coeffcients using
a simple and lightweight architecture. In our setup, the random encoding $z$ is a 100-D vector. The generator and discriminator are both fully connected neural networks consisting of 4 layers each, with 100 nodes in each layer.
Each layer is followed by a batch normalization step.
Following the guidelines of previous architectures \cite{wu2016learning}, our discriminator
uses a LeakyReLU activation while our generator uses regular ReLU.

The discriminator is trained by minimizing the vanilla GAN loss described as follows:
\begin{equation}\label{eqn:gan}
	\mathcal{L}_d = \mathbb{E}_{x\sim{\cal T}} [ \log \left(D(x)\right) ] + \mathbb{E}_{z\sim U} [ \log \left(1-D(G(z))\right) ].
\end{equation}
where $x$ represents the shape coefficients, $D$ is the discriminator, $G$ is the generator, $U$ represents an uniform distribution of real numbers in $(-1, 1)$,
and $\mathcal{T}$ is the training data.
In our experiments, we noticed that using the traditional loss for the generator leads to a highly
unstable training where the generated data converges to a single mode (which loses diversity).
To overcome this issue, we employ an approach similar to the one proposed in \cite{improvedGAN}.
Specifically, let $f(x)$ be the intermediate activations of the discriminator given an input $x$.
Our generator will try to generate samples that match some statistics of the activations of
the real data, namely mean and covariance.
Thus, the generator loss is defined as follows:
\begin{equation}\label{eqn:generator}
	\mathcal{L}_g = \norm{\mathbb{E}_{x\sim{\cal T}} [ f(x) ] - \mathbb{E}_{z\sim U} [ f(G(z)) ]}_2^2 +
					\norm{cov_{x\sim{\cal T}} [ f(x) ] - cov_{z\sim U} [ f(G(z)) ]}_2^2
\end{equation}
where $cov$ is the vectorized covariance matrix of the activations.
Using this loss results in a much more stable learning procedure.
During all our experiments the single mode problem never occurred, even when
training the GAN for thousands of epochs.
We use the Adam optimizer~\cite{Adam} with a learning rate of $10^{-4}$ for the discriminator and $0.0025$ for the
generator.
Similarly to \cite{wu2016learning}, we only train the discriminator if its accuracy is below 80\%.

\begin{figure}[t]
	\includegraphics[width=1.0\linewidth]{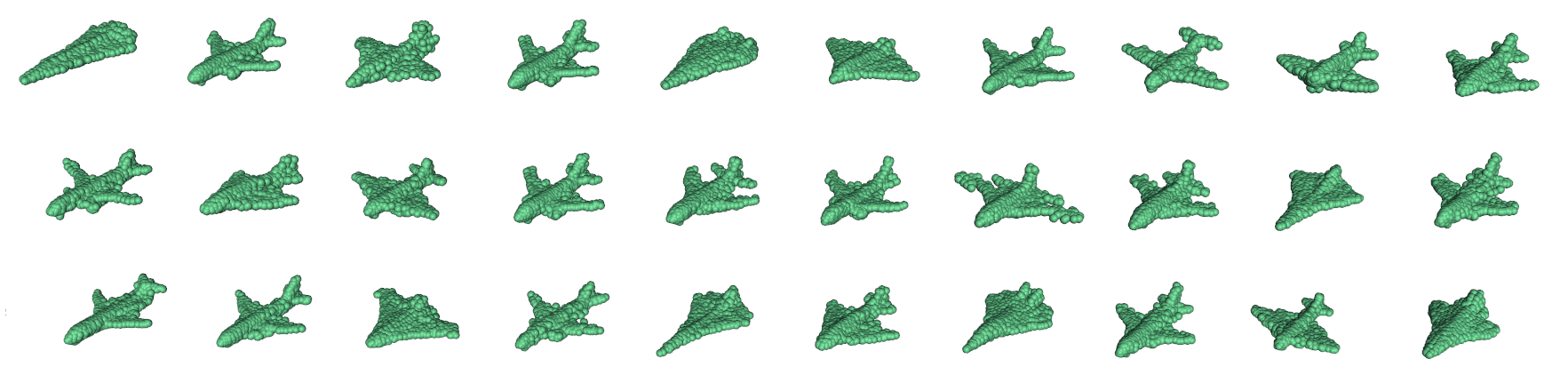}
	\includegraphics[width=1.0\linewidth]{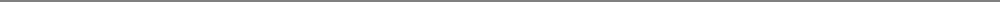}
	\includegraphics[width=1.0\linewidth]{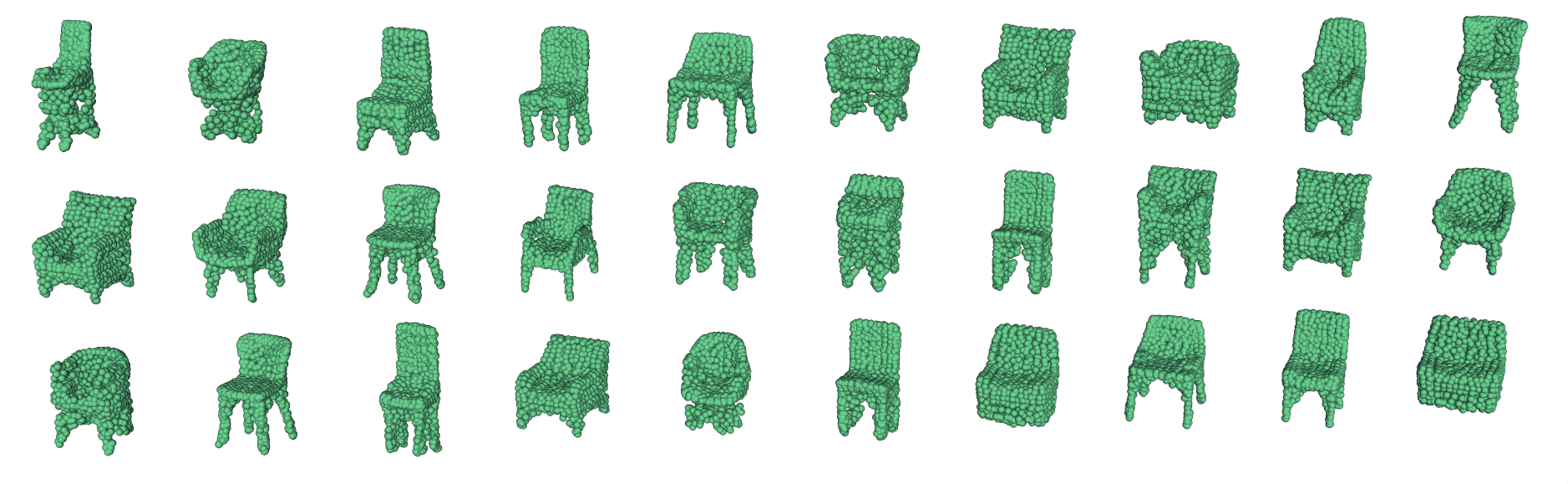}
	\includegraphics[width=1.0\linewidth]{images/hline.png}
	\includegraphics[width=1.0\linewidth]{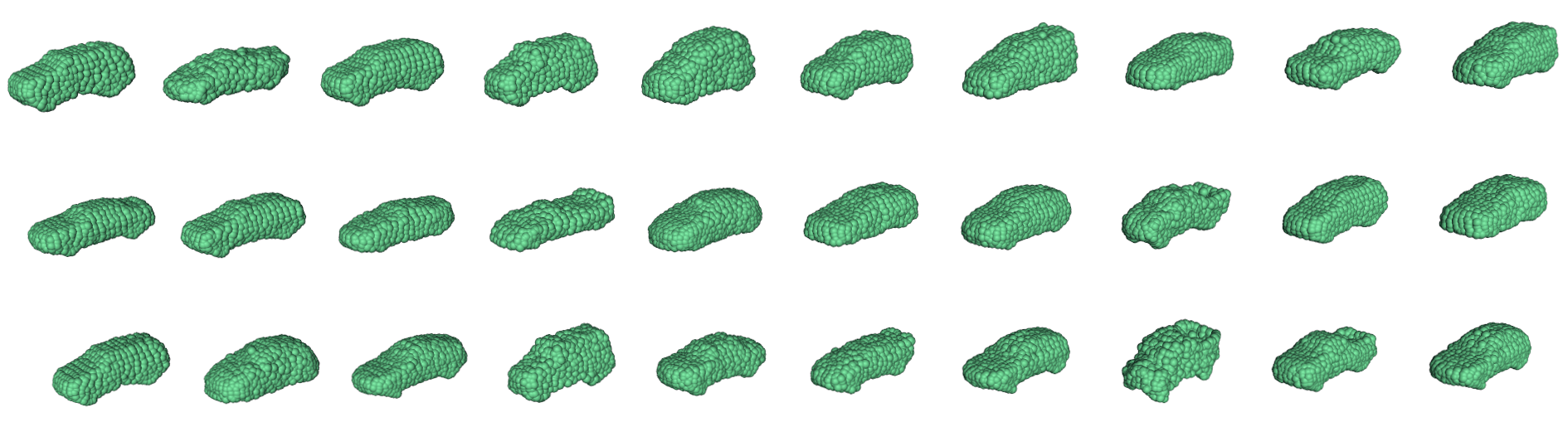}
	\vspace{-16pt}
	\caption{\small \label{fig:gallery} A gallery showing results of using our method to generate points clouds for three categories: airplane, chair, and car. We use our method to train a GAN for each category separately. The training is generally very fast and completes within a few minutes. The results shown here are generated by randomly sampling the encoding $z$ of the GAN.}
	\vspace{-12pt}
\end{figure}

\begin{figure}[t]
	\includegraphics[width=1.0\linewidth]{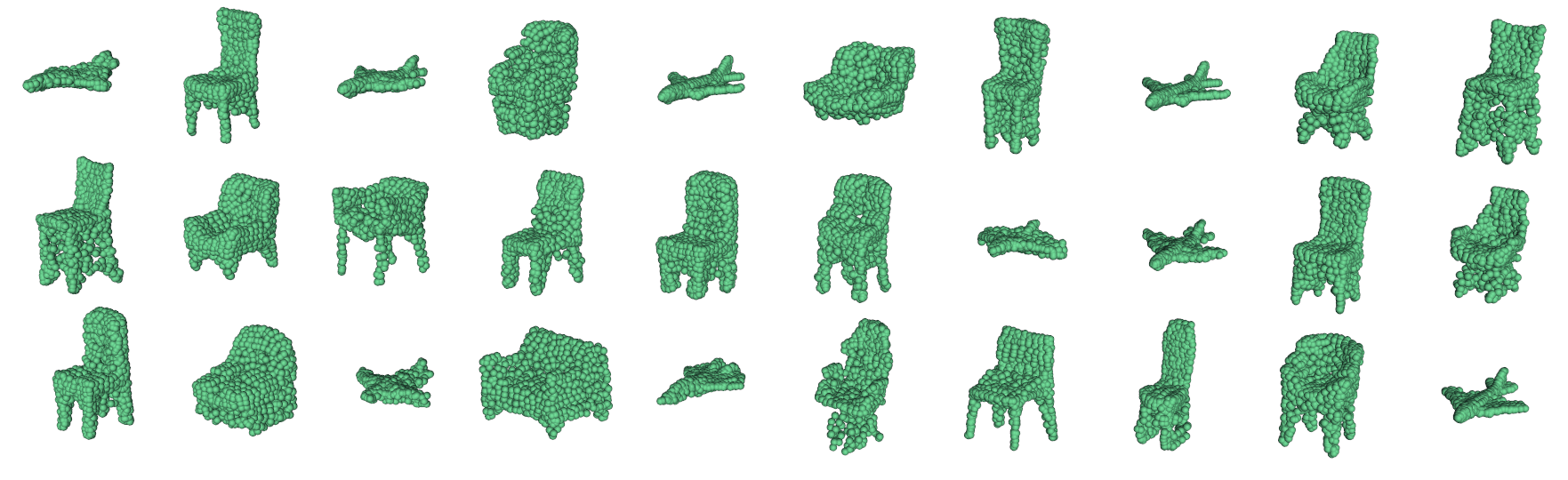}
	\vspace{-16pt}
	\caption{\small \label{fig:mixed} Results for a mixed category (chair + airplane) showing the ability of our method to capture multi-modal distributions over mixed-category shapes.}
	\vspace{-6pt}
\end{figure}

\begin{figure}[t]
\includegraphics[width=1.0\linewidth]{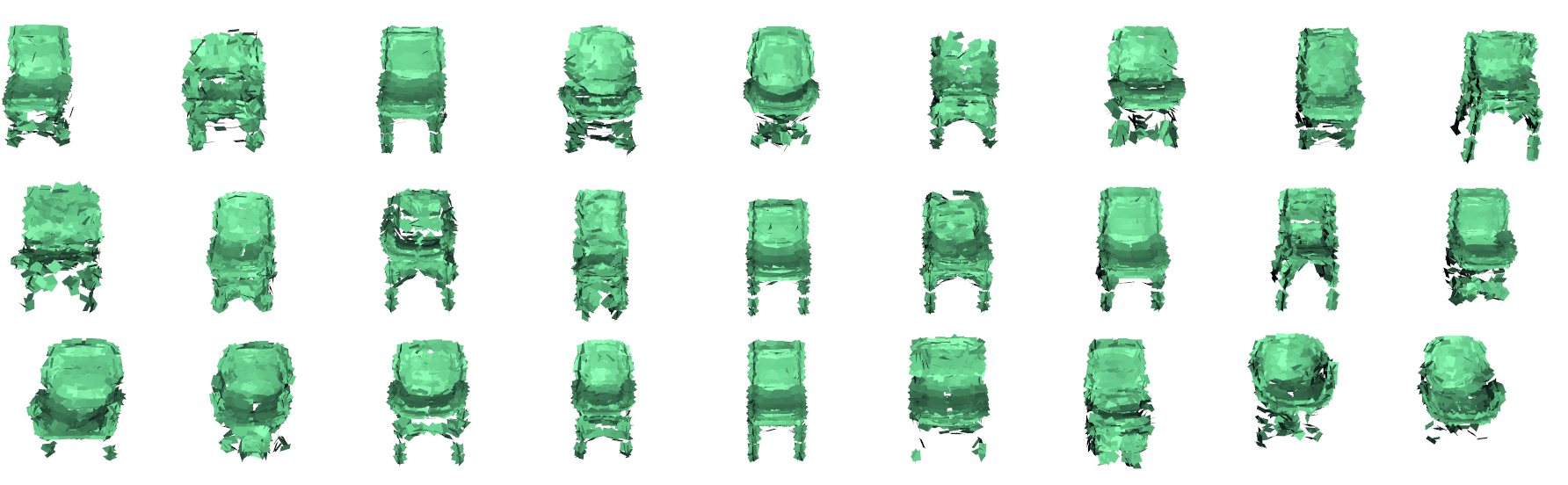}
\vspace{-16pt}
\caption{\small \label{fig:normals} Chairs generated with normal. For visualization we shade each point as a square patch centered at the point and oriented by the normal. This shows the ability of our method to generate not only $x,y,z$ coordinates but also incorporate associated point attributes such as normal.}
\vspace{-12pt}
\end{figure}

\section{Experiments}\label{sec:experiments}

\noindent \textbf{Training data.} To generate training data, we use several shape categories from the ShapeNet dataset~\cite{chang2015shapenet}, including chairs, airplanes, cars etc. We sample each shape with 1K Poisson disk sample points using the algorithm described in~\cite{Bowers:2010:PPD}. Poisson disk samples evenly disperse the points over the surface, which turns out to work better at preserving geometric details than using white noise samples. We can easily increase the number of sample points to 4K or 8K and beyond. Unlike voxel-based representations, our method is lightweight, and increasing the sample size only leads to moderate increases in computation resources and time. 


\vspace{12pt}
\noindent \textbf{Qualitative evaluation.} Figure~\ref{fig:gallery} shows a gallery of results generated using our method for each of the three categories: airplane, chair, and car. The results are generated by randomly sampling the encoding $z$ and demonstrate a variety of shapes within each category. The training is very fast and generally completes within a few minutes. This is an order of magnitude faster than training deep neural networks built upon voxel representations. Figure~\ref{fig:mixed} shows additional results for a mixed category that combines shapes from the chair and airplane datasets. For this mixed category we used $B=300$ basis. The results show the ability of our method to capture the multi-modal distributions over mixed-category shapes.

\vspace{12pt}
\noindent \textbf{Generating multiple point attributes.} Our method can generate points with multiple attributes, such as surface normal, color, by simply appending these attributes to the $(x,y,z)$ coordinates. The overall procedure remains the same except the shape basis is learned over the joint space of positions and normals etc. Figure~\ref{fig:normals} shows chair results generated with normal. The ability to incorporate point attributes is an additional advantage over voxel-based representations (which do not explicitly represent surface information of shapes).


\begin{table}[t]
\centering
\begin{tabular}{c|cccc|c}
	Dataset & GAN(10) & GAN(50) &	GAN(100) &	SGAN (100) & PPCA (100) \\
\hline
	Chairs		& 2.57	& 2.53	& 2.37	&\textbf{2.19} &2.88 \\
	Airplanes	& 1.96	& 1.93	& 1.94	&\textbf{1.48} &2.29 \\
	Cars		& 1.45	& 1.42	& 1.44	&\textbf{1.25} &1.59 \\
	Tables		& 2.88	& 2.68	& 2.66	&\textbf{2.34} &3.18 \\
\end{tabular}
\caption{\small \label{tab:quant} Distance (Eq.\ref{eq:chamfer-distance}) between the generated samples and training samples for different generative models. The numbers in parentheses indicate the number of PCA coefficients used for each column. SGAN is the GAN trained using the sorted data. The GAN approach outperforms the PPCA baseline by a considerable margin even without thesorting procedure.}
\vspace{-6pt}
\end{table}

\begin{figure}[t]
\includegraphics[width=1.0\linewidth]{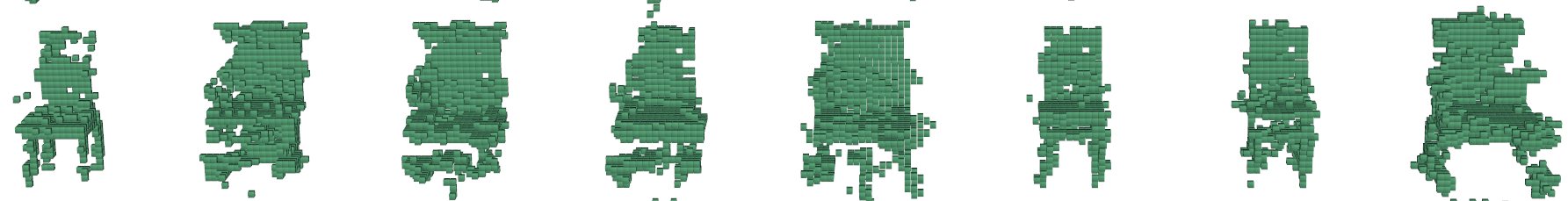}
\vspace{-16pt}
\caption{\small \label{fig:3dgan} 3D-GAN result for the chair category. The models are generated by following~\cite{wu2016learning}.}
\vspace{-12pt}
\end{figure}

\begin{figure}[t]
\centering
\includegraphics[width=0.8\linewidth]{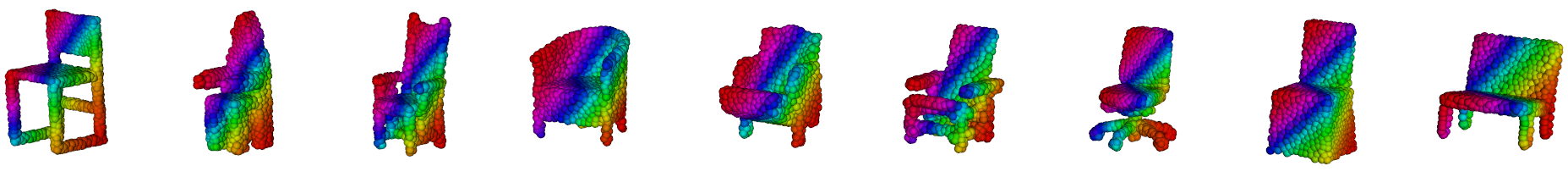}
\includegraphics[width=0.8\linewidth]{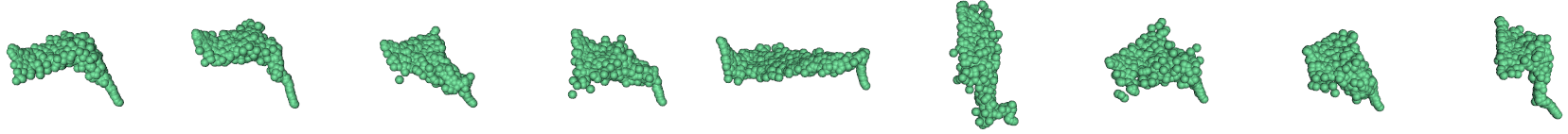}
\vspace{-8pt}
\caption{\small \label{fig:chairsxyz} Sorting point clouds using $x+y+z$ values. Top row shows a visualization of the training data using this sorting strategy.
	Bottom row shows the generated shapes for the chair category. They are visually of poor quality compared to kd-tree sorting.}
\vspace{-6pt}	
\end{figure}

\begin{figure}[t]
\centering
\includegraphics[width=0.8\linewidth]{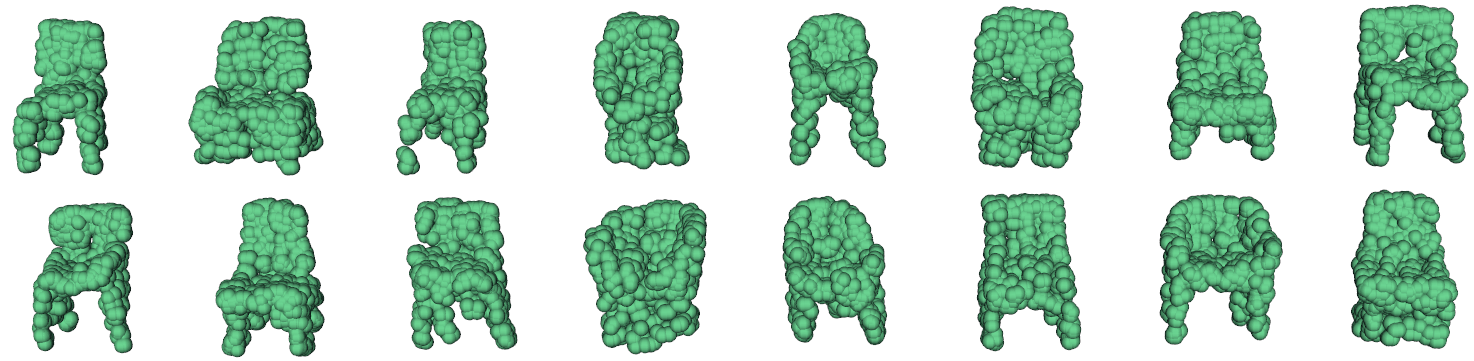}
\caption{\small \label{fig:1dconv} Samples from an alternative GAN architecture using 1D convolutions. Trained using the the point clouds directly.}
\vspace{-12pt}
\end{figure}

\vspace{12pt}
\noindent \textbf{Quantitative evaluation.}
We compare variations of our model to a PPCA baseline~\cite{tipping1999probabilistic}. The PPCA model performs a linear factor analysis of the data using: $y \sim \mathbf{W}x + \mu + \sigma$. The matrix $\mathbf{W}$ is a basis, the latent variables $x \sim N(0, I)$, noise $\sigma \sim N(0, \sigma^2I)$ and the $\mu$ is the data mean. In other words, PPCA learns an independent Gaussian distribution over the coefficients $x$, whereas our approach employs a GAN. We compare PPCA results with variations of our model by changing the number of basis and examining its influence on the quality of the results.
The metric used in the evaluation is defined as follows.
Let $\mathcal{T}$ and $\mathcal{S}$ be the set of training and generated samples, respectively.
We define our distance measure $d(\mathcal{T}, \mathcal{S})$ using a variant of the Chamfer distance, as follows:
\begin{equation}\label{eq:chamfer-distance}
	d(\mathcal{T}, \mathcal{S}) = 
	\frac{1}{|\mathcal{T}|} \sum_{t \in \mathcal{T}} \min_{s \in \mathcal{S}} \norm{t - s}_2 +
	\frac{1}{|\mathcal{S}|} \sum_{s \in \mathcal{S}} \min_{t \in \mathcal{T}} \norm{t - s}_2
\end{equation}
The results can be seen in Table~\ref{tab:quant}. Our approach that uses a GAN to model the distribution of coefficients consistently outperforms the PPCA baseline, which models the distribution as a Gaussian. For the chairs and tables categories the difference between the PPCA and GAN is large, suggesting that the distribution of the coefficients is highly multi-modal. The results by varying the number of bases are also shown in the Table~\ref{tab:quant}. Increasing the number of basis beyond a hundred did not improve our results further.

\vspace{12pt}
\noindent \textbf{Visual comparison to 3D-GAN.} To compare our results with the 3D-GAN model~\cite{wu2016learning}, we followed their description to implement our own version of the algorithm as there is no publicly available implementation that can be trained from scratch. Figure~\ref{fig:3dgan} shows the 3D-GAN results for the chair category. As in~\cite{wu2016learning}, the training data is generated by voxelizing each shape to $64^3$ resolution, and we employ the same hyper-parameters for our GAN model as theirs. Our results, which can be found in Figure~\ref{fig:gallery}, compare favorably to 3D-GAN. In addition, our network is significantly smaller and faster to train.


\vspace{12pt}
\noindent \textbf{The role of the \emph{kd-tree}.} The kd-tree induces a shape-dependent but consistent ordering of points across shapes. Moreover the ordering is locality preserving, i.e., two points that are close in the underlying 3D shape are also likely to be close in the list after kd-tree ordering. We believe that this property is critical for the estimating a good initial basis for the shape representation. In order to verify this hypothesis we consider an alternative scheme where the points are ordered according to their $x+y+z$ value. Although consistent across shapes this ordering does not preserve locality of the points and indeed yields poor results as seen in Figure~\ref{fig:chairsxyz}. However, other data structures that preserve locality such as locality-sensitive hashing~\cite{gionis1999similarity} and random-projection trees~\cite{dasgupta2008random} are possible alternatives to kd-trees.

We also experimented an scheme for generating shapes where \emph{1D~convolutions} on the ordered points are used for both the generative and discriminative models in a GAN framework. Instead of learning a linear shape basis with has wide support over all the points, the 1D-GAN architecture only has local support. Since the ordering is locality sensitive, one might expect that convolutional filters with small support are sufficient for generation and discrimination. This approach can also be robust to a partial reordering of the list due to variations in the shape structures. Moreover, the 1D-GAN can be directly learned on the ordered point list without having to first learn a bases, and is even more compact than the GAN+PCA basis approach. The architecture used for this experiments has the same number of layers with our standard approach. The major difference is in the fact that we use 1D convolutional layers instead of fully connected ones. The generator layers have a filter size of 25 and the first one has 32 filters. The following layers double the number filters of the previous layer. The discriminator is the mirrored version of the generator. Figure~\ref{fig:1dconv} shows the results obtained using the 1D-GAN for the chair category. Remarkably, the generated shapes are plausible, but are ultimately of worse quality than our GAN+PCA approach. Both these experiments suggest that the kd-tree plays a important role for our method.

\begin{figure}[h]
  \includegraphics[width=1.0\linewidth]{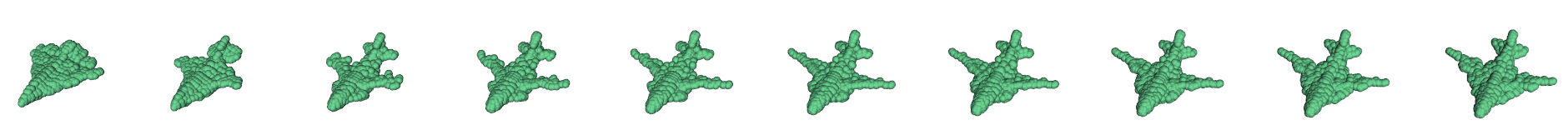}
  \includegraphics[width=1.0\linewidth]{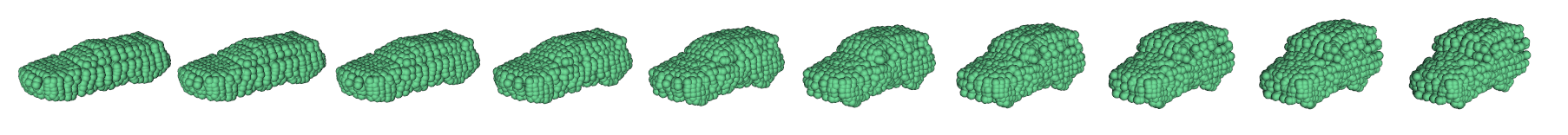}
  \includegraphics[width=1.0\linewidth]{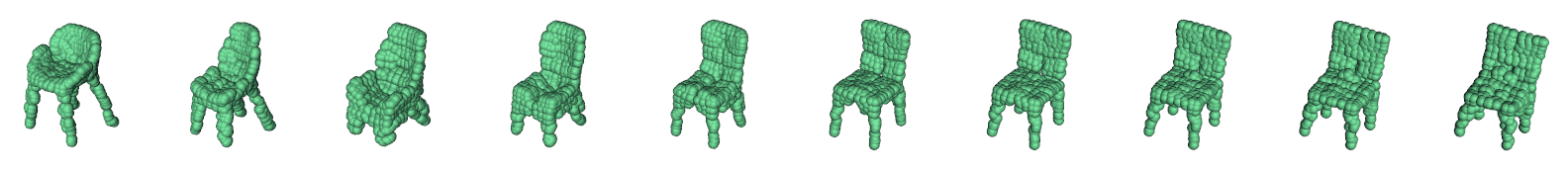}
\vspace{-16pt}
\caption{\small \label{fig:interpolation} Interpolation of the encodings $z$ between a start shape and an end shape for each of the three categories shown here: airplane, car, and chair.}
\vspace{-12pt}
\end{figure}

\vspace{12pt}
\noindent \textbf{Shape interpolation.} Similar to image-based GAN and 3D-GAN, we can perform shape interpolation by linearly interpolating in the encoding space $z$. Specifically, we can pick two encodings $z_1$, $z_2$, linearly interpolate them, and use our generative model to compute the resulting point cloud. The interpolation results are shown in Figure~\ref{fig:interpolation}. As observed, the interpolated shapes are plausible and exhibit non-linearity that cannot be achieved by directly interpolating the shape coefficients.

\section{Conclusion and Future Work} \label{sec:conclusion}
We showed that conventional CNN architectures can be used to generate 3D shapes as point clouds once they are ordered using kd-trees.
We found that a hundred linear basis are generally sufficient to model a category of diverse shapes such as chairs.
By employing GANs to model the multi-modal distribution of the basis coefficients we showed that our method outperforms the PPCA baseline approach.
The ordering of points produced by the kd-tree also allows reasonable shape generation using 1D-GANs.
Our approach is of comparable quality but considerably more lightweight than 3D voxel-based shape generators.
Moreover it allows the incorporation of multiple point attributes such normals and color in a seamless manner.
In future work we aim to explore if improving the point orderings during training improves the reconstruction error.
We also plan to investigate the role of space-partitioning data structures on 3D shape classification and segmentation tasks.

\paragraph{Acknowledgement:}
This research was supported in part by the NSF grants IIS-1617917, IIS-1423082 and ABI-1661259, and a faculty gift from Facebook. The experiments were performed using high performance 
computing equipment obtained under a grant from the Collaborative R\&D Fund 
managed by the Massachusetts Tech Collaborative.

\bibliography{egbib}
\end{document}